\documentclass[sigconf]{acmart}

\setcopyright{none}
\makeatletter
\def\@copyrightpermission{}
\makeatother

\AtBeginDocument{%
  \providecommand\BibTeX{{%
    \normalfont B\kern-0.5em{\scshape i\kern-0.25em b}\kern-0.8em\TeX}}}

\acmSubmissionID{73}

\usepackage{amssymb}
\usepackage{algorithm}
\usepackage{algorithmic}

\definecolor{citecolor}{RGB}{119,185,0} 

\usepackage{pifont}
\usepackage{color}
\DeclareMathOperator*{\argmax}{argmax}

\newlength\savewidth

\usepackage[font=small, labelfont=bf, textfont=normalfont]{caption}

\begin{document}

\title{Hybrid, Unified and Iterative: A Novel Framework for Text-based Person Anomaly Retrieval}


\author{Tien-Huy Nguyen$^*$}
\affiliation{%
  \institution{University of Information Technology \\
  Vietnam National University}
  \city{Ho Chi Minh}
  \country{Vietnam}
}

\author{Huu-Loc Tran$^*$}
\affiliation{%
  \institution{University of Information Technology \\
  Vietnam National University}
  \city{Ho Chi Minh}
  \country{Vietnam}
}

\author{Huu-Phong Phan-Nguyen$^*$}
\affiliation{%
  \institution{University of Information Technology \\
  Vietnam National University}
  \city{Ho Chi Minh}
  \country{Vietnam}
}

\author{Quang-Vinh Dinh}
\affiliation{%
  \institution{AI VIETNAM Lab}
  \city{Ninh Thuan}
  \country{Vietnam}
}


\renewcommand{\shortauthors}{Tien-Huy Nguyen, Huu-Loc Tran, Huu-Phong Phan-Nguyen, \&  Quang-Vinh Dinh}

\thanks{* All authors contributed equally to this paper. \\ 
This research is partly supported by AI VIETNAM \cite{aivietnamVit}.}


\begin{abstract}
Text-based person anomaly retrieval has emerged as a challenging task, with most existing approaches relying on complex deep-learning techniques. This raises a research question: How can the model be optimized to achieve greater fine-grained features? To address this, we propose a Local-Global Hybrid Perspective (LHP) module integrated with a Vision-Language Model (VLM), designed to explore the effectiveness of incorporating both fine-grained features alongside coarse-grained features. Additionally, we investigate a Unified Image-Text  (UIT) model that combines multiple objective loss functions, including Image-Text Contrastive (ITC), Image-Text Matching (ITM), Masked Language Modeling (MLM), and Masked Image Modeling (MIM) loss. Beyond this, we propose a novel iterative ensemble strategy, by combining iteratively instead of using model results simultaneously like other ensemble methods. To take advantage of the superior performance of the LHP model, we introduce a novel feature selection algorithm based on its guidance, which helps improve the model's performance. Extensive experiments demonstrate the effectiveness of our method in achieving state-of-the-art (SOTA) performance on PAB dataset, compared with previous work, with a 9.70\% improvement in R@1, 1.77\% improvement in R@5, and 1.01\% improvement in R@10.
\end{abstract}

\copyrightyear{2025}
\acmYear{2025}
\setcopyright{acmlicensed}\acmConference[WWW Companion '25]{Companion
Proceedings of the ACM Web Conference 2025}{April 28-May 2, 2025}{Sydney, NSW, Australia}
\acmBooktitle{Companion Proceedings of the ACM Web Conference 2025 (WWW Companion '25), April 28-May 2, 2025, Sydney, NSW, Australia}
\acmDOI{10.1145/3701716.3717653}
\acmISBN{979-8-4007-1331-6/2025/04}

\ccsdesc[500]{Information systems ~ Text-based image retrieval}

\keywords{Multimedia Retrieval, Deep Learning, Representation Learning.}

\maketitle


\section{Introduction}

Text-based person retrieval, a well-established task \cite{yang2023unifiedtextbasedpersonretrieval, li2024dataaugmentationtextbasedperson, Zheng_2020}, involves retrieving specific indicators from large-scale image databases using textual queries. The current method exhibits biases toward common actions and only solves their actions, limiting diversity and the generalizability of models, particularly for detecting abnormal behaviors. To address this, we explore text-based anomaly person retrieval, leveraging advancements in deep learning and computer vision \cite{maprach,mmm2023,tienhuyaic, nguyen2024improvinggeneralizationvisualreasoning, googleGoogleDrive,googleGoogleDrive2}. Unlike traditional methods that use entire images, we focus on localized image regions to enhance attention to fine details. Furthermore, we adopt multiple objective learning to model complex image-text features and propose a novel iterative ensemble strategy to iteratively refine model predictions, capitalizing on the strengths of multiple models. Therefore, our primary contributions are:


\begin{itemize}
    \item A hybrid approach blending local and global perspectives (LHP), enhancing the model's ability to utilize both fine-grained and holistic visual information.

    \item Unified Image-Text (UIT) Modeling integrates MIM \cite{simmim}, MLM \cite{mlmloss}, ITC \cite{itcloss} and ITM \cite{itmloss} tasks, leveraging LHP-based feature selection for efficient and accurate multi-modal representation learning.

    \item  We introduce a novel iterative ensemble algorithm that utilizes the results of multiple models more effectively, helping to improve the overall performance.

    \item Our comprehensive experiments demonstrate the effectiveness of our proposed method, achieving SOTA performance in text-based person anomaly retrieval on real-world datasets.

\end{itemize}

\section{Method}
\begin{figure*}
    \centering
    \includegraphics[width=0.6\linewidth]{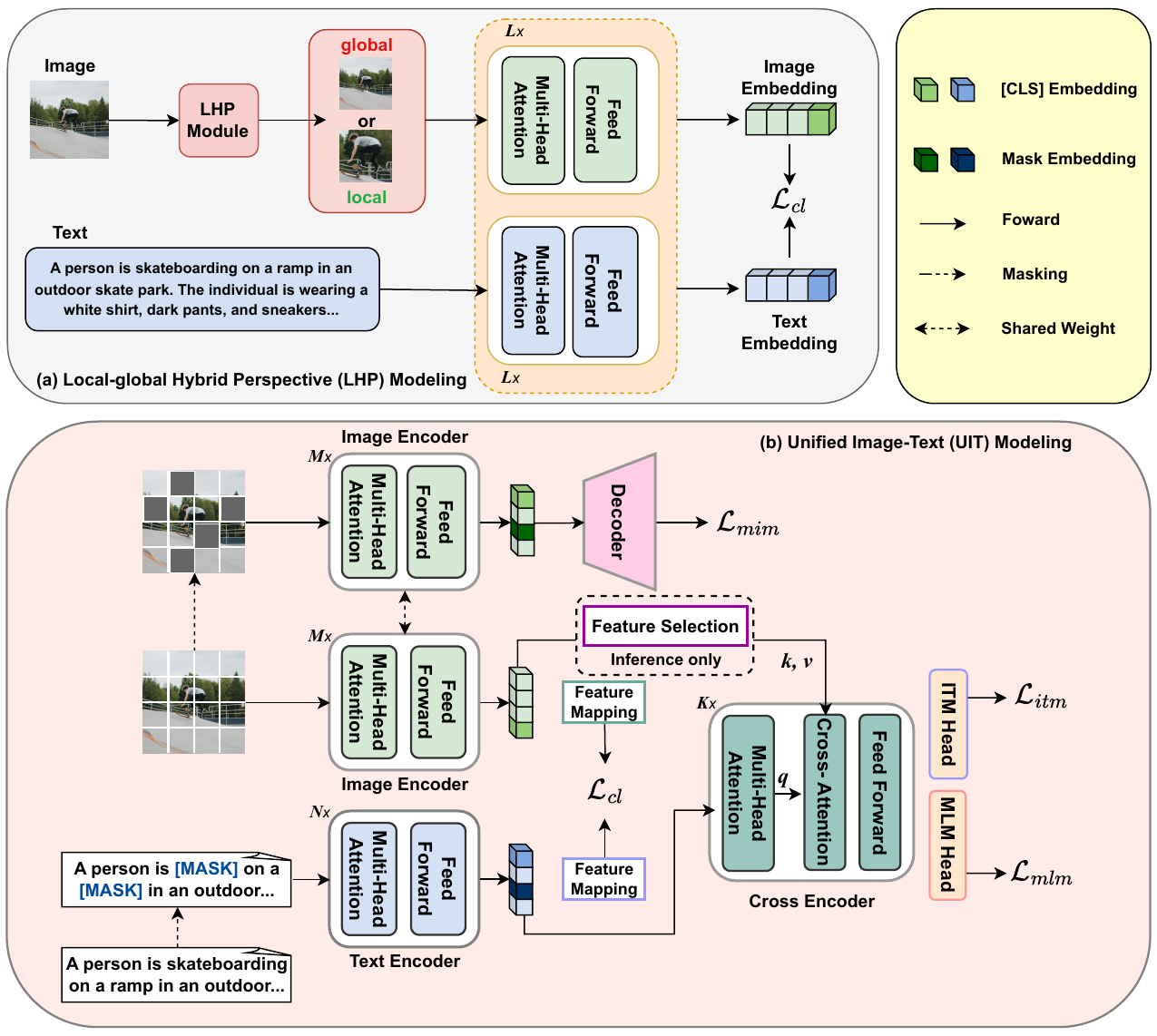}
    \vspace{-14em}
    \caption{
    \textbf{(a) Overview of Local-global Hybrid Perspective (LHP) Modeling.} It processes an image probabilistically, applying either a local transform for fine-grained details or a global transform for comprehensive context. Contrastive learning aligns image and text embeddings by minimizing distances for matching pairs and maximizing distances for non-matching pairs. \textbf{(b) Unified Image-Text (UIT) Modeling with Feature Selection.} UIT is a cross-modal framework that integrates MIM, MLM, ITC, and ITM to unify image and text understanding. UIT enhances inference by leveraging LHP-based feature selection for efficient and accurate multi-modal representation learning.
    }
    \vspace{-1.5em}
    \label{fig:overview}
\end{figure*}

\subsection{Local-global Hybrid Perspective Modeling}
\label{section:lhp}
In this section, we introduce LHP modeling, as illustrated in \textbf{\autoref{fig:overview}(a)}. The LHP module determines whether to process an image locally or globally based on a probabilistic criterion. The LHP module takes an input image I, along with a local transform and a global transform. A random value is sampled from a normal distribution with a mean of 0.5 and a variance of $1\div6$. If the sampled value is greater than 0.5, the local transform is applied to image I; otherwise, the global transform is applied. The module then outputs either the locally transformed or globally transformed image I, based on the sampled value.





    



In the local perspective, a region of interest is cropped from the image to focus on fine-grained details. In contrast, the global perspective retains the entire image to capture holistic contextual information. This hybrid approach allows the model to benefit from both granular and comprehensive views of the image.



To align the image and text embeddings, we employ a contrastive loss function that minimizes the distance between embeddings of matching pairs while maximizing the distance between embeddings of non-matching pairs. For a given image-text pair(I,T), their feature representations are extracted as follows: 
\vspace{-0.6em} 
\begin{equation}
f_i = \mathcal{E}_i(I), \space f_t = \mathcal{E}_t(T).
\end{equation}
where $\mathcal{E}_i, \mathcal{E}_t$ is the Image Encoder, Text Encoder.

The image-to-text similarity $S_{\text{I2T}}$ within the
batch is defined as follows:
\begin{equation}
S_{\text{I2T}} = \frac{\exp\left(s(f_i, f_t)/\tau\right)}{\sum_{j=1}^N \exp\left(s(f_i, f_t^j)/\tau\right)} 
\end{equation}
where $s(\cdot, \cdot)$ is the cosine similarity, $\tau$ is a temperature parameter. Finally, the contrastive learning loss is presented below:

\vspace{-1em}
\begin{equation}
\mathcal{L}_{\text{cl}} = -\frac{1}{2} \mathbb{E}\left[\log S_{\text{I2T}} + \log S_{\text{T2I}}\right] 
\end{equation}


\subsection{Unified Image-Text (UIT) Modeling}
\label{sec:mtl}

\subsubsection{Overall framework}

In this section, we introduce UIT, as illustrated in \textbf{\autoref{fig:overview}(b)}, a cross-modal model developed for our work. Drawing inspiration from CMP \cite{yang2024walkinglargescaleimagetextbenchmark}, UIT enhances its ability to learn robust visual representations by incorporating masked image reconstruction into its training process.

UIT comprises four main components: Image Encoder ($\mathcal{E}_i$), Text Encoder ($\mathcal{E}_t$), Cross Encoder ($\mathcal{E}_{cross}$) and a Decoder ($\mathcal{D}$). Given an image ($I$) -text ($T$) pair, Random Masking is applied to generate a masked image ($I_{masked}$) and text ($T_{masked}$). The masked image passes through $\mathcal{E}_i$ and $\mathcal{D}$ for the Masked Image Modeling (MIM) task. Meanwhile, $I$ and $T_{masked}$ are encoded by $\mathcal{E}_i$ and $\mathcal{E}_t$ for the Image-Text Contrastive (ITC) task, producing image embeddings ($f_i$) and text embeddings ($f_t$), respectively. These embeddings are combined in the Cross Encoder ($\mathcal{E}_{cross}$) to get cross embeddings ($f_{cross}$), which are utilized for the Image-Text Matching (ITM) and Masked Language Modeling (MLM) tasks.

\textbf{Masked Image Modeling (MIM).} The MIM aims to reconstruct masked image patches for improving the visual representations of $\mathcal{E}_i$. Given an input image $I$, the image is divided into patches, and a subset of patches is randomly masked to get the masked image $I_{masked}$. The model is trained to predict these masked patches using unmasked ones. The objective of \textbf{MIM} is computed as:
\vspace{-0.3em} 
\begin{equation}
    \hat{I} = \mathcal{D}(\mathcal{E}_i(I_{masked})),
\end{equation}
\vspace{-1.3em} 
\begin{equation}
    \mathcal{L}_{mim} = \frac{1}{N} \sum_{k=1}^N \| \hat{I}_i - I_i \|_1,
\end{equation}
\vspace{-0.3em} 
where $N$ is the number of images, $\hat{I}_i$ is the i-th reconstructing image, and $I_i$ is the original image.

\textbf{Image-Text Contrastive (ITC) Learning.} The goal of contrastive learning is to maximize the similarity between positive pairs while minimizing the similarity between negative pairs. The contrastive loss is computed as described in \textbf{Section~\ref{section:lhp}}.


\textbf{Image-Text Matching (ITM) Learning:} The Image-Text Matching determines whether a given image-text pair corresponds. The cross-encoder processes the text embeddings as input and integrates the image embeddings using a cross-attention mechanism at each layer. The output of the cross encoder is projected into a 2-dimensional space using \textit{ITM head} (a Multi-Layer Perceptron (MLP)). The Image-Text Matching loss is then defined as:
\vspace{-0.2em} 
\begin{equation}
\mathcal{L}_{itm} = -\mathbb{E} \Big[ p(I, T) \log \hat{p}(I, T) + \big(1 - p(I, T)\big) \log \big(1 - \hat{p}(I, T)\big) \Big],
\end{equation}

where \(p(I, T) \in \{0, 1\}\) is the ground truth label for whether the image-text pair matches (\(1\) for matching, \(0\) for non-matching), \(\hat{p}(I, T)\) is the predicted probability of the pair matching.

\textbf{Masked Language Modeling (MLM):} The MLM aims to reconstruct masked text tokens from the context. The masked text \( T_{masked} \), along with the corresponding person image \( I \), is fed into the cross encoder. The cross-encoder processes these inputs and outputs a fused representation, which is passed through an \textit{MLM head}. The objective of MLM learning is to predict the likelihood of the masked token \( t \) in \( T_{masked} \), given the image \( I \) and the unmasked parts of the text. The training process minimizes the cross-entropy loss:
\vspace{-0.2em} 
\begin{equation}
    \mathcal{L}_{mlm} = -\mathbb{E}\Big[p_{\text{mask}}(I, T_{masked}) \log \hat{p}_{\text{mask}}(I, T_{masked})\Big],
\end{equation}

where \( \hat{p}_{\text{mask}}(I, T_{masked}) \) is the predicted likelihood of the masked token \( t \) in \( T_{masked} \), \( p_{\text{mask}}(I, T_{masked}) \) is the ground truth one-hot vector representing the correct token.

Given the above optimization objectives, the full training loss is formulated as:
\vspace{-0.2em} 
\begin{equation}
    \mathcal{L} = \mathcal{L}_{itc} + \mathcal{L}_{itm} + \mathcal{L}_{mlm} + \alpha \mathcal{L}_{mim},
\end{equation}
where \(\alpha\) denotes the weight for the MIM loss and we set $\alpha = 0.1356$.

\subsubsection{Feature Selection}

\begin{algorithm}
\caption{Feature Selection Algorithm}
\label{alg:feature_selection}
\begin{algorithmic}[1]
\STATE \textbf{Input:} Image embeddings $f_i$, LHP similarity matrix \textit{sim\_matrix}  

\STATE \textbf{Output:} \textit{topk} image features for each text embedding.
\STATE $\textit{selected\_features} \gets []$
\FOR{$index, row$ \textbf{in enumerate(}\textit{sim\_matrix})}
    \STATE $\textit{topk\_sim}, \textit{topk\_idx} \gets \textit{topk}(\textit{row})$
    \STATE $\textit{selected\_features}[index] \gets f_i[\textit{topk\_idx}]$
\ENDFOR
\RETURN \textit{selected\_features}
\end{algorithmic}
\end{algorithm}

During the inference stage, we leverage the similarity matrix obtained from the LHP model to select the \textit{top-k} image features with the highest similarity scores, as detailed in \textbf{Algorithm~\autoref{alg:feature_selection}}. This approach takes advantage of the superior performance of the LHP model, which is more effective at identifying high-quality top-\(k\) candidates compared to the similarity matrix computed directly by the UIT model. The similarity matrix is calculated based on the cosine similarity between \(f_i\) and \(f_t\).

By using the LHP model for feature selection, the process focuses on the most relevant image features, enabling the model to prioritize important images. The selected \textit{top-k} image features and their corresponding text features are then fed into the cross-encoder. The cross-encoder processes these features and computes the final matching scores via the ITM Head, ensuring accurate alignment between image and text modalities.

\subsection{Iterative Ensemble}

Ensemble learning is a powerful technique that combines the predictions of multiple models. By aggregating outputs from diverse models, ensemble methods can reduce overfitting and enhance generalization. To utilize that, we propose a novel method for ensemble, named as iterative ensemble, described as in \textbf{Algorithm~\autoref{alg:ie}}.

\label{sec:ie}
\begin{algorithm}
\caption{Iterative Ensemble Algorithm}
\label{alg:ie}
\begin{algorithmic}[1]
\STATE \textbf{Input:} Gallery set $I$, queries $Q$, model list $\theta$ , ground truth $gt$, 

\hspace{1cm} scoring function $f_s$, weight value set $W \in [0,1]$
\STATE \textbf{Output:} $\text{Best score matrix S.}$

\STATE $\text{Initialize} \ S\gets 0$

\FOR{$i, t_\theta$ \textbf{in enumerate}($\theta$)}

    \STATE $\text{pred} \gets \text{topk}(w \cdot S + (1-w) \cdot t_\theta(I, Q), \ \text{dim} = 1)$
    \STATE $w \gets \argmax\limits_{\substack{argmax \\ w \in W}}(f_s(\text{pred}, gt))$

    \STATE $S \gets w \cdot S + (1-w) \cdot t_\theta(I, Q)$

\ENDFOR

\RETURN $\text{Best score matrix S.}$ 
\end{algorithmic}
\end{algorithm}

where $pred \in \mathbb{R}^{n \times n}$ with $n$ is equal to number of images. In addition, to be able to choose the weight for the best results, we perform hyperparameter tuning with W and we realized that when setting w values close to the value 1, the results improve significantly.  

Most of the current ensemble methods mostly use simultaneous model predictions. The difference between our method compared to others is that it retains most of the scores of the current score, and \textbf{gradually} references the scores of the next models. Initially, S is equal to 0, and after the first iteration, S receives the  of the first model. However, in the next iterations, S will be combined between the results of the previous model (multiplied by $w$) with the results of the current model (multiplied by $(1-w)$). Finally, we obtain the final S while still ensuring references from different models, allowing if the result is agreed upon by many models, the score is still high, and vice versa. Furthermore, when the results are uncertain, an iterative ensemble acts as a reranker, helping to change rankings. 

\section{Experimental Results}
\subsection{Implementation in details}
We integrate the (LHP) module into BEiT-3, baseline model, \cite{beit3} to combine local and global features, and use Swin-B and a BERT-based encoder for unified image-text modeling (UIT). During inference, we apply an iterative ensemble strategy with BEiT-3 \cite{beit3}, UIT, BLIP-2 \cite{blip2}, and CLIP \cite{open-clip} to enhance performance. BEiT-3 with LHP is fine-tuned for 3 epochs (batch size 184, image size 384×384), and UIT for 22 epochs (batch size 84, image size 224×224), Both experiments use cosine annealing for learning rate scheduling, the AdamW optimizer, and initialized initial learning rate value is $10^{-5}$.

\subsection{Dataset and Evaluation Metrics}
\textbf{The Pedestrian Anomaly Behavior} (PAB) \cite{yang2024walkinglargescaleimagetextbenchmark} dataset includes 1,013,605 synthesized and 1,978 real-world image-text pairs, covering diverse actions and anomalies. Real-world videos provide the test data, while a diffusion model generates the training set. Performance is evaluated using recall rates (R@K). A search is successful if the image perfectly matches the text query that appears in the top k-ranked images. Higher R@K values indicate better performance, and results are reported for R@1, R@5, and R@10.

\subsection{Quantitative Results}


We evaluate our proposed method against APTM and CMP on the Text-based Person Anomaly retrieval, with results shown in \textbf{\autoref{tab:sota_comparison}}.

\begin{table}[H]
\centering
\begin{tabular}{lccc}
\hline
\textbf{Method}                & \textbf{R@1}   & \textbf{R@5}   & \textbf{R@10}  \\ \hline
\multicolumn{4}{c}{\textbf{0.1M} training images} \\ \hline
APTM\cite{yang2023unifiedtextbasedpersonretrieval} 0shot           & 9.40          & 22.14         & 30.18         \\ 
APTM\cite{yang2023unifiedtextbasedpersonretrieval}  tuned       & 69.92         & 95.60         & 97.32         \\ 
IHNM\cite{yang2024walkinglargescaleimagetextbenchmark}                   & 72.25         & 95.91         & 98.03         \\ 
PE \ \ \ \ \ \cite{yang2024walkinglargescaleimagetextbenchmark}  & 71.79         & 95.40         & 97.83         \\ 
CMP \ \cite{yang2024walkinglargescaleimagetextbenchmark}       & 72.80         & 96.01         & 97.47         \\ 
\textbf{Ours}                    & \textbf{85.39} & \textbf{99.49} & \textbf{99.95} \\ \hline
\multicolumn{4}{c}{\textbf{1M} training images} \\ \hline
CMP \ \cite{yang2024walkinglargescaleimagetextbenchmark} & 79.53         & 97.93         & 98.84         \\ 
\textbf{Ours} & \textbf{89.23} & \textbf{99.70} & \textbf{99.85} \\ \hline
\end{tabular}
\caption{Comparison with other methods on 0.1M and 1M training images of PAB dataset.}
\label{tab:sota_comparison}
\end{table}
\vspace{-1.3em} 
 Our method achieves SOTA performance across all metrics (R@1, R@5, and R@10) on the PAB dataset. When trained on 0.1M images, our method outperforms CMP, the second-best approach, with a substantial 12.59\% improvement in Recall@1, achieving 85.39\%. On the full dataset of 1M images, our method maintains its advantage, improving Recall@1 by 9.70\%, reaching 89.23\%. These results demonstrate the robustness and scalability of our approach, which outperforms existing methods across varying dataset sizes.


\subsection{Ablation Study}

As described in \textbf{\autoref{tab:table2}}, The only-global method achieves high R@10 (99.95\%) but lower R@1. The only-local approach slightly improves R@1 (85.39\%) by emphasizing localized details but underperforms slightly at R@10 (99.90\%). It is easy to see that the local-global hybrid perspective (LHP) gives superior results, which proves that combining both will help the model give the best and most stable results, instead of using only one of them.

\begin{table}[H]
\centering
\renewcommand{\arraystretch}{1.2}
\begin{tabular}{lccc}
\toprule
\textbf{Method}      & \textbf{R@1} & \textbf{R@5} & \textbf{R@10} \\ \midrule
only-global          & 85.24            & 99.44            & 99.95            \\ 
only-local           & 85.39            & 99.44            & 99.90            \\ 
LHP (ours)           & \textbf{85.39}   & \textbf{99.49}   & \textbf{99.95}   \\ \bottomrule
\end{tabular}
\caption{Ablation experiments for LHP model. Only-global, only-local, and LHP use cropped, entire images respectively, and use both as model input.}
\label{tab:table2}
\end{table}
\vspace{-1em}

As shown in the \textbf{\autoref{tab:table3}
}, the baseline model trained on 0.1M images, R@1 is 85.24\%. Adding LHP improves R@1 to 85.39\%, a 0.18\% increase, showing the benefit of leveraging local-global hybrid perspectives.

\begin{table}[h]

\centering
\renewcommand{\arraystretch}{1.2}
\resizebox{8.5cm}{!}{\
\begin{tabular}{lcccc}
\toprule
\textbf{Method}                         & \textbf{Training Images} & \textbf{R@1} & \textbf{R@5} & \textbf{R@10} \\ \midrule
Baseline                                & 0.1M                     & 85.24             & 99.44             & \textbf{99.95}             \\
Baseline + LHP                          & 0.1M                     & 85.39             & 99.49             & \textbf{99.95}             \\
Baseline + LHP                          & 1M                       & 87.11             & 99.65             & 99.85             \\
Baseline + LHP + UIT (FS)                & 1M                       & 88.37             & 99.70             & 99.85             \\
Baseline + LHP + UIT (FS) + IE         & 1M                       & \textbf{89.23}    & \textbf{99.70}    & 99.85   \\ \bottomrule
\end{tabular}
}
\caption{Ablation experiments for highlighting the progressive improvements in person anomaly retrieval performance achieved by incorporating LHP, UIT (FS: feature selection), and (IE: iterative ensemble).}
\label{tab:table3}
\end{table}

 When scaling to 1M images, LHP further boosts R@1 to 87.11\%, representing a significant 1.72\% improvement over the 0.1M training setup. Incorporating UIT with feature selection (FS) elevates R@1 to 88.37\%, a 1.26\% gain compared to using LHP alone on 1M images, demonstrating the effectiveness of multi-modal objectives. Finally, adding iterative ensemble (IE) alongside LHP and UIT (FS) achieves the highest R@1 of 89.23\%, a 0.86\% increase over the previous setup and a total 3.99\% improvement over the baseline. 

\begin{table}[H]
\centering

\resizebox{8.5cm}{!}{%
\begin{tabular}{ccccccc}
\toprule
\textbf{Attempts} & \textbf{Iter.1 (UIT)} & \textbf{Iter.2 (BLIP-2)} & \textbf{Iter.3 (CLIP)} & \textbf{R@1} & \textbf{R@5} & \textbf{R@10} \\
\midrule
1  & 0      & -      & -      & 87.11 & 99.65 & 99.85 \\
2  & 0.5    & -      & -      & 84.07 & 99.14 & 99.85 \\
3  & 0.8    & -      & -      & 87.26 & 99.70  & 99.85 \\
4  & 0.85   & -      & -      & 87.41 & 99.70  & 99.85 \\
5  & 0.875  & -      & -      & 87.97 & 99.70  & 99.85 \\
6  & 0.9    & -      & -      & 88.22 & 99.70  & 99.85 \\
7  & 0.9125 & -      & -      & 88.27 & \textbf{99.75} & 99.85 \\
8  & 0.925  & -      & -      & 88.37 & 99.70  & 99.85 \\
9  & 0.9375 & -      & -      & 88.17 & 99.70  & 99.85 \\
10 & 0.95   & -      & -      & 87.97 & 99.70  & 99.85 \\
11 & 0.925  & 0.875  & -      & 88.83 & 99.70  & 99.85 \\
12 & 0.925  & 0.9    & -      & 88.88 & \textbf{99.75} & \textbf{99.90}  \\
13 & 0.925  & 0.925  & -      & 88.68 & 99.70  & 99.85 \\
14 & 0.925  & 0.9    & 0.85   & 88.88 & 99.70  & 99.85 \\
15 & 0.925  & 0.9    & 0.8725 & \textbf{89.23} & 99.70  & 99.85 \\
16 & 0.925  & 0.9    & 0.9    & \textbf{89.23} & 99.70  & 99.85  \\
17 & 0.925  & 0.9    & 0.9125 & 89.18 & 99.70  & 99.85  \\
18 & 0.925  & 0.9    & 0.925  & 89.13 & 99.70  & 99.85  \\
\bottomrule
\end{tabular}%
}
\caption{Iterative ensemble results with varying weights for subsequent iterations and their impact on R@1, R@5, and R@10 performance.}
\label{tab:iterative_dataset}
\end{table}
\vspace{-1em}

As presented in \textbf{\autoref{tab:iterative_dataset}}, analyze the performance of iterative ensembles using various configurations of the UIT, BLIP-2, and CLIP models. The iterative adjustments in the UIT model (Iter.1) show steady improvements in retrieval performance, with Recall@1 increasing from an initial setting of 0 to values like 0.85 and 0.925, while Recall@5 and Recall@10 remain consistently high at 99.70\% and 99.85\%, respectively. The inclusion of BLIP-2 (Iter.2) and CLIP (Iter.3) further enhances performance, reaching a Recall@1 peak of 89.23\% in later iterations (e.g., Iter.1 = 0.925, Iter.2 = 0.9, Iter.3 = 0.9). This study demonstrates the effectiveness of iterative optimization and the synergy of combining complementary models to achieve state-of-the-art retrieval accuracy.

\vspace{-1.6em}

\section{Conclusion}
In this paper, we proposed a novel framework for Text-based Person Anomaly Retrieval by introducing the LHP and UIT modeling. LHP effectively combines fine-grained local details with global contextual information, while UIT integrates multiple loss objectives, including MIM, MLM, ITC, and ITM. Furthermore, we introduced a novel feature selection algorithm and an iterative ensemble strategy, which significantly enhance the retrieval performance by leveraging complementary models and refining predictions.

Extensive experiments on the PAB dataset demonstrate that our method achieves SOTA performance, with improvements over previous methods. Specifically, our approach achieves a 9.7\% improvement in Recall@1 on the 1M full dataset and a 12.59\% improvement on the 0.1M dataset. Ablation studies further validate the effectiveness of each component, showing the advantages of combining local and global perspectives and the impact of feature selection and ensemble strategies.

{\small
\bibliographystyle{ACM-Reference-Format}
\bibliography{egbib}
}

\end{document}